\documentclass[10pt,twocolumn,letterpaper]{article}

\usepackage{cvpr}
\usepackage{times}
\usepackage{graphicx}
\usepackage{amsmath}
\usepackage{amssymb}
\usepackage{booktabs}
\usepackage{graphics} 
\usepackage{epsfig} 
\usepackage{float}
\usepackage{cite}
\usepackage{multirow}
\usepackage{svg}
\usepackage{algorithm}
\usepackage{algpseudocode}
\usepackage{graphicx}
\usepackage{import}
\usepackage{subcaption}

\usepackage[pagebackref=true,breaklinks=true,letterpaper=true,colorlinks,bookmarks=false]{hyperref}

\usepackage[capitalize]{cleveref}
\crefname{section}{Sec.}{Secs.}
\Crefname{section}{Section}{Sections}
\Crefname{table}{Table}{Tables}
\crefname{table}{Tab.}{Tabs.}

\begin{document}

\title{Monocular 3D Object Detection with LiDAR Guided Semi Supervised Active Learning}

\author{Aral Hekimoglu\\
Technical University Munich\\
Munich, Germany\\
{\tt\small aral.hekimoglu@tum.de}
\and
Michael Schmidt\\
BMW Group\\
Munich, Germany\\
{\tt\small michael.se.schmidt@bmw.de}
\and
Alvaro Marcos-Ramiro\\
BMW Group\\
Munich, Germany\\
{\tt\small alvaro.marcos-ramiro@bmw.de}
}

\maketitle

\begin{abstract}

We propose a novel semi-supervised active learning (SSAL) framework for monocular 3D object detection with LiDAR guidance (MonoLiG), which leverages all modalities of collected data during model development. We utilize LiDAR to guide the data selection and training of monocular 3D detectors without introducing any overhead in the inference phase. During training, we leverage the LiDAR teacher, monocular student cross-modal framework from semi-supervised learning to distill information from unlabeled data as pseudo-labels. To handle the differences in sensor characteristics, we propose a data noise-based weighting mechanism to reduce the effect of propagating noise from LiDAR modality to monocular. For selecting which samples to label to improve the model performance, we propose a sensor consistency-based selection score that is also coherent with the training objective. Extensive experimental results on KITTI and Waymo datasets verify the effectiveness of our proposed framework. In particular, our selection strategy consistently outperforms state-of-the-art active learning baselines, yielding up to 17\% better saving rate in labeling costs. Our training strategy attains the top place in KITTI 3D and bird’s-eye-view (BEV) monocular object detection official benchmarks by improving the BEV Average Precision (AP) by 2.02.

\end{abstract}

\section{Introduction} \label{sec:introduction}

3D object detection is fundamental in scene understanding for autonomous driving vehicles. Detectors operating on point cloud scans from the LiDAR sensor achieve impressive performance on benchmarks like KITTI \cite{geiger2012we}; however, they are costly for consumer vehicles. Monocular RGB cameras offer a cheaper alternative. Therefore, there has been a surge of interest in research on monocular 3D object detectors. Convolutional Neural Network (CNN) based monocular detectors achieve state-of-the-art (SOTA) performance with the help of massive annotated datasets. However, annotating a large amount of 3D detection data is time and labor-consuming. Specifically for monocular 3D object detectors, manually annotating 3D boxes from monocular imagery is infeasible due to a lack of depth information. Therefore, LiDAR point clouds are recorded during data collection, and annotators label 3D box locations on the collected point clouds. To save annotation costs, only the most informative frames in the collected samples are labeled to train models. Consequently, large amounts of LiDAR data with beneficial 3D information remain unlabeled.

\begin{figure}[t]
    \centering
    \includegraphics[width=1.0\linewidth]{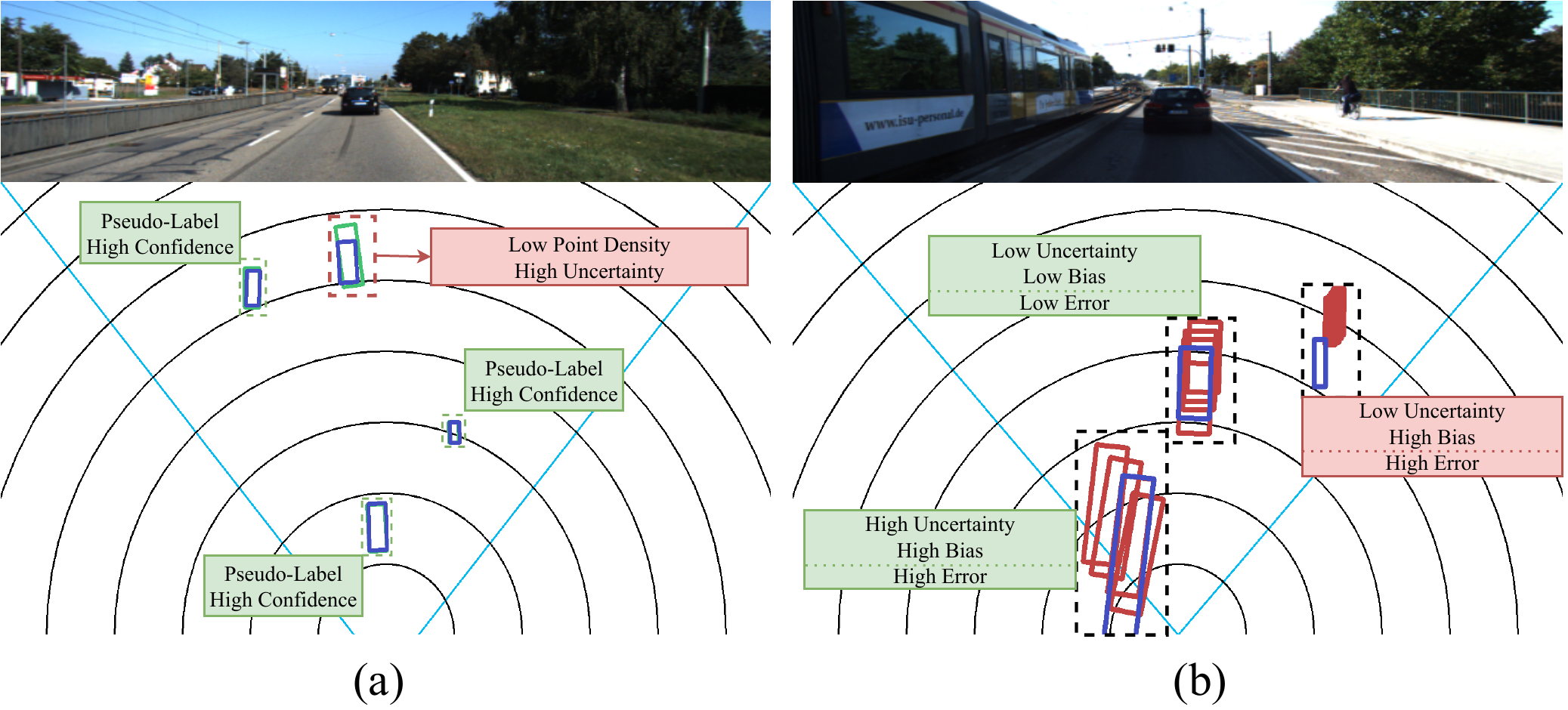}
    \caption{RGB image and predictions in BEV space of 2 frames from KITTI. (a) In regions with low point cloud return, the LiDAR detector’s predictions (green) are not safe to use as pseudo-labels since they do not perfectly overlap with ground truth (blue). (b) In predictions with low uncertainty but high variation from the ground-truth (bias), uncertainty from an ensemble of monocular detectors (red) is not enough to capture erroneous samples.}
    \label{fig:intro}
\end{figure}

Semi-supervised learning (SSL) and active learning (AL) are two related techniques that aim to improve model performance while minimizing labeling effort by utilizing unlabeled data. AL focuses on selecting the most informative samples for labeling, while SSL focuses on training the model using unlabeled data.

In a recent cross-modal SSL method \cite{peng2021lidar}, predictions from the LiDAR detector (teacher) are treated as the ground truth of unlabeled data. They are combined with annotations of labeled data to train the monocular detector (student). However, as shown in \cref{fig:intro}a, some predictions from the LiDAR detector are not accurate, and thus learning from them is not optimal for the monocular detector. We observe that these inaccurate predictions are typically from areas with low point cloud densities, i.e., distant or occluded objects. These correspond to regions in LiDAR where aleatoric (data) uncertainty is high \cite{feng2021review}.

The key idea of AL for object detection is to leverage the current detector to select the most informative samples for labeling under a fixed-labeling budget. The selection is based on an acquisition function that estimates the detector’s uncertainty on samples. Then, samples with high uncertainty are selected for labeling. However, as shown in \cref{fig:intro}b, some detections with low uncertainty are still far from the ground truth. For these samples, having an acquisition function that measures this discrepancy is essential.

To solve the problems mentioned above, we present the MonoLiG framework, illustrated in \cref{fig:intro-ssal}, that consists of a coherent selection and training phase. During our training phase (\cref{sec:method-semisupervised}), we utilize the cross-modal teacher-student framework with a LiDAR teacher and a monocular student detector. To reduce the effect of incorrect LiDAR predictions, we propose to scale the loss of the monocular detector based on the confidence of generated labels. To this end, we extend the LiDAR detector with an aleatoric uncertainty estimation head and define the confidence of predicted labels with the aleatoric uncertainty of LiDAR. During our selection phase (\cref{sec:method-activelearning}), inspired by the cross-modal teacher-student frameworks from SSL, we extend the uncertainty-based selection score and use LiDAR predictions as pseudo-labels to measure the distance of monocular predictions to the ground truth. To our knowledge, our work is the first to leverage the teacher-student paradigm for AL selection and integrate it with a coherent SSL training strategy. By combining AL and SSL, MonoLiG is able to select challenging samples that are difficult to learn with semi-supervised training and thus achieve higher model performance with minimal labeling costs.

Our main contributions are summarized as follows:
\begin{itemize}
    \item We propose MonoLiG, a novel framework that consists of a coherent selection and training phase. The proposed strategies outperform AL and SSL baselines separately, and achieves the best performance when utilized coherently.
    \item We extend current uncertainty strategies for AL selection by adapting the teacher-student paradigm and adding an inconsistency term, resulting in a better data savings rate than the SOTA AL baselines.
    \item We identify the potential error propagation from the teacher to the student model in cross-modal teacher-student SSL methods and propose a pseudo-label weighting mechanism based on the aleatoric uncertainty of the teacher. Our proposed training strategy define a new SOTA for monocular 3D object detection in the KITTI test benchmark.
\end{itemize}

\section{Related work} \label{sec:related-work}
\subsection{Active learning for object detection} \label{sec:rel-al}

Pool-based AL selection methods can be grouped into two categories: uncertainty-based \cite{beluch2018power, choi2021active, gal2017deep} and diversity-based \cite{sener2018active, agarwal2020contextual, sinha2019variational}. One approach to estimate the uncertainty is through ensembles \cite{beluch2018power}. Different models are trained with different random initializations to construct a committee with slightly different predictions for uncertain samples. Then, the informativeness score is obtained by an acquisition function like entropy \cite{shannon1948mathematical}, or BALD \cite{houlsby2011bayesian} for classification tasks or total variance (TV) \cite{tsymbalov2018dropout} for regression tasks. In contrast, diversity-based methods target maintaining the distribution of the unlabeled pool by selecting a set of samples that covers the remaining points within a distance. Core-set \cite{sener2018active} uses Euclidean distance in the feature space learned by CNNs, and CDAL \cite{agarwal2020contextual} utilizes KL-divergence between context features, which they define as a mixture of predicted softmax probabilities in a detection network. One recent task-agnostic approach, LL4AL \cite{yoo2019learning}, trains a loss-learning module during training and uses the predicted loss as the score to select samples.

AL is extended for 2D object detection \cite{elezi2022not, haussmann2020scalable, yu2022consistency, aghdam2019active}, and 3D object detection from LiDAR \cite{choi2021active, schmidt2020advanced, feng2019deep}. Elezi \etal \cite{elezi2022not} select samples using uncertainty and robustness of the detector, defined by the consistency between a sample and its augmented version. Yu \etal \cite{yu2022consistency} propose a 2-stage selection strategy. First, they select samples using a consistency-based metric and continue selection with a score that promotes the class distribution of selected samples to be different from the labeled pool.

Most works for AL for object detection focus on classification \cite{haussmann2020scalable, yuan2021multiple, aghdam2019active} and ignore localization of the bounding boxes. Choi \etal \cite{choi2021active} estimate the aleatoric and epistemic uncertainty for both classification and localization and combine the uncertainties in a single selection score. In \cite{schmidt2020advanced}, Schmidt \etal train an ensemble of models and define the localization uncertainty as the intersection over union (IoU)-based matching score. Since localization is more challenging for monocular 3D detectors, we also utilize a localization uncertainty-based selection score and extend it with a criterion to capture the deviation of ensemble predictions from the ground-truth.

\subsection{Semi-supervised learning in object detection}\label{sec:rel-ss}

SSL aims to improve the performance of a model by training with a limited labeled dataset and exploiting information from a large amount of unlabeled data. SSL approaches can be categorized into two groups: consistency regularization \cite{jeong2019consistency, laine2017temporal, chen2020temporal} and pseudo-labeling \cite{lee2013pseudo, caine2021pseudo, mugnai2021soft}. Consistency regularization trains the model's parameters on unlabeled data by penalizing the inconsistency between predictions for the same input under different perturbations. In pseudo-labeling methods, \cite{lee2013pseudo}, a trained model predicts labels for unlabeled samples. Then the model is updated by training on the unlabeled data using the \textit{pseudo-labels} as the optimization target.

An issue with pseudo-labeling is overfitting to incorrect predictions due to the confirmation bias \cite{arazo2020pseudo}. One solution is filtering pseudo-labels based on a confidence score (hard-thresholding). FixMatch \cite{sohn2020fixmatch} enhances the quality of pseudo-labels by filtering predictions from the teacher with low classification confidence. Wang \etal \cite{wang20213dioumatch} extend this approach to 3D object detection by using an additional IoU-based localization confidence score. Another solution to confirmation bias is using soft-pseudo-labels\cite{rizve2021defense, shi2018transductive} and scaling the effect of each prediction based on their confidence.

Recently, the teacher-student paradigm has been used for monocular 3D object detection in a cross-modal setting to transfer information from one modality (LiDAR) to another (monocular) \cite{peng2021lidar, chong2022monodistill}. Chong \etal \cite{chong2022monodistill} distill information from the LiDAR detector with feature and label guidance. Similar to our work, Peng \etal \cite{peng2021lidar} generate pseudo-labels using a LiDAR teacher model to train a student monocular detector. We extend their approach by using the aleatoric uncertainty of LiDAR as the confidence score of pseudo-labels.

\subsection{Semi-supervised active learning}\label{sec:rel-ssal}

Recent works \cite{wang2020semi, huang2021semi, song2019combining ,simeoni2021rethinking, gao2020consistency} combine SSL and AL using semi-supervised techniques like pseudo-labeling in the training phases of AL cycles to distill information from the unlabeled data. Huang \etal \cite{huang2021semi} constructs a Mean Teacher \cite{tarvainen2017mean} by applying an exponential moving average (EMA) to weights obtained at the end of each AL cycle. They extend pool-based AL with training from unlabeled samples using the predictions from the mean teacher. Gao et al. \cite{gao2020consistency} proposed an AL framework that utilizes a selection score based on augmentation consistency with a SSL training strategy penalizing augmentation inconsistency.

\section{Methodology} \label{sec:methods}

\subsection{Optimization problem} \label{sec:method-optimization}

\begin{figure}[t]
    \centering
    \includegraphics[width=1.0\linewidth]{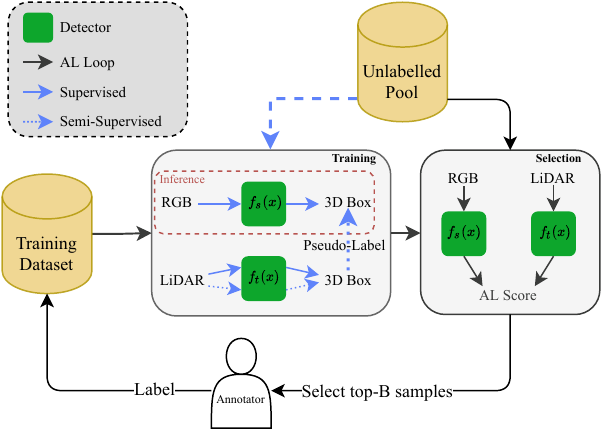}
    \caption{Overview of the proposed MonoLiG framework. At each cycle, we first train our LiDAR detector with the training dataset and the monocular detector in a semi-supervised manner with LiDAR predictions as pseudo-labels. During selection, we use predictions from both detectors to compute a selection score for each unlabeled sample. Then, an annotator labels samples with highest score under a budget $B$.}
    \label{fig:intro-ssal}
\end{figure}

\begin{figure*}[t]
    \centering
    \includegraphics[width=0.98\linewidth]{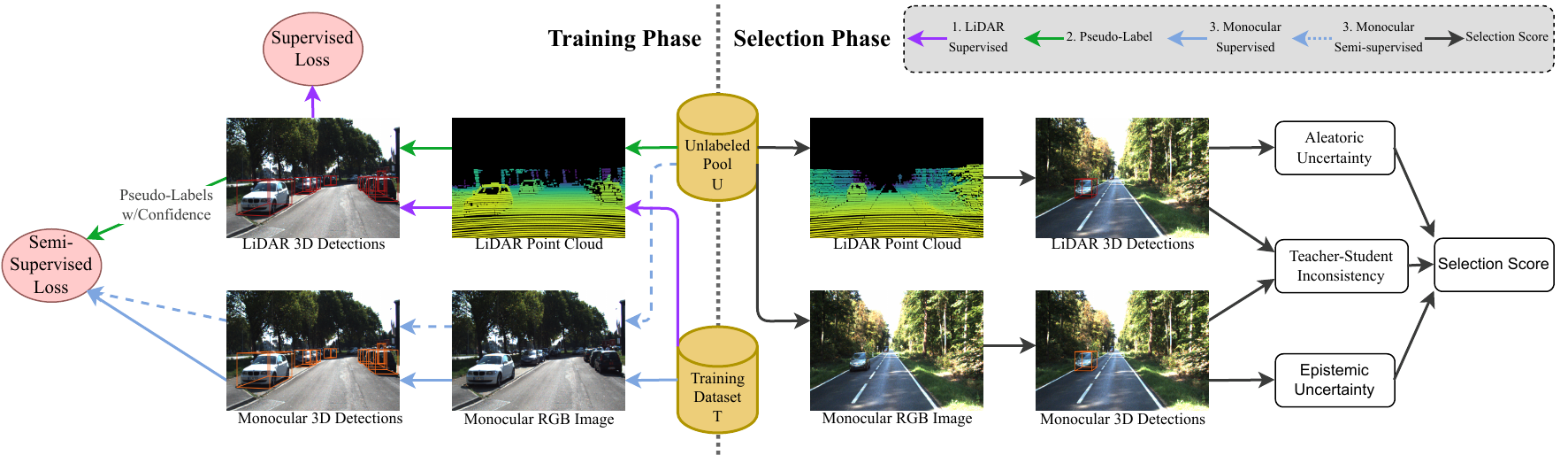}
    \caption{Illustration of the proposed MonoLiG framework. \textbf{In the training phase}, 1) We train a LiDAR detector as our teacher model. 2) Using its predictions on unlabeled data, we generate pseudo-labels and assign confidence based on the aleatoric uncertainty. 3) We train a monocular detector as our student model with SSL. \textbf{In the selection phase}, we select samples based on the epistemic uncertainty of the student model, the inconsistency between predictions from the teacher and the student, and the aleatoric uncertainty of the teacher.}
    \label{fig:overview}
\end{figure*}

Let $(x,y)$ be a sample pair drawn from the dataset space $D$. For an input scene $x$ - consisting of a synchronized point cloud and an image - label $y$ contains bounding box parameters $b_o$ and a semantic class label $c_o$ for all objects $o$ within the scene. We use the cross-modal teacher-student paradigm \cite{peng2021lidar}, where the student model $f_s$ is a monocular detector and the teacher model $f_t$ is a LiDAR detector. During training, we use both models, for inference, we only deploy the student model. Therefore, the optimization objective is to reduce the expected loss of the student model $f_s(x;\theta)$ given by,
\begin{equation} \label{eqn:optimization_integral}
E[L;\theta] = \iint_{x,y \sim D}{L(f_s(x_s;\theta),y) dxdy}
\end{equation}
where $L(f_s(x;\theta),y)$ represents a loss function. We formulate our theory using mean-square error (MSE) as the regression loss for a single bounding box and compute the sample loss as the sum of losses of all boxes. We decompose \cref{eqn:optimization_integral} into two components (derivation in the supplement):
\begin{equation} \label{eqn:loss+alea}
L(f_s(x;\theta),y) = (f_s(x;\theta)-h_s(x))^2 + (h_s(x)-y)^2
\end{equation}

where $h_s(x)$ is the global optimum for $f_s(x;\theta)$. The first component of this equation can be optimized during training, whereas the second component represents $u^{al}_{s}$, the aleatoric uncertainty of the student (data noise), that cannot be reduced by optimization.

We follow pool-based AL, where we assume access to a labeled sample set $(x_T , y_T)$ belonging to the training dataset $T$, and to unlabeled samples $(x_U)$ from a large unlabeled data pool $U$, randomly (i.i.d.) sampled from the dataset space $D$. Then, our optimization objective is approximated as follows:
\begin{multline} \label{eq:optimization}
E[L;\theta] \approx \sum_{(x_T,y_T) \epsilon {T}}{ (f_s(x_T;\theta)-y_T)^2}\\
+ \sum_{(x_U,y_U) \epsilon {U}}{ (f_s(x_U;\theta)-h_s(x_U))^2 + (h_s(x_U)-y_U)^2 }
\end{multline}

\subsection{MonoLiG overview}

The MonoLiG framework, illustrated in \cref{fig:intro-ssal}, optimizes \cref{eq:optimization} during each AL cycle through two phases: a semi-supervised training phase and an active learning selection phase. During the training phase (\cref{sec:method-semisupervised}), we train with supervised learning using the labels $y_T$ as our target. We employ the teacher-student paradigm on unlabeled data and use the predictions from the teacher model $f_t(x)$ as a proxy target for $f_s(x)$ in the form of pseudo-labels. We extend this paradigm with a pseudo-labeling weighting mechanism based on aleatoric uncertainty. During the selection phase (\cref{sec:method-activelearning}), we minimize the expected loss of the student model by selecting a subset $S$ from $U$, under a fixed budget $|S| = B$, to be labeled by an oracle and moved to $T$ for retraining the student model in the next cycle. We propose a scoring function consisting of three components: epistemic uncertainty of the student, inconsistency between the predictions of the teacher and the student, and aleatoric uncertainty of the teacher to be coherent with our training phase. The training and selection cycle repeats until a stop condition is satisfied, i.e., the detector’s performance converges for several iterations or reaches the desired performance.

\subsection{Training phase with semi-supervised learning} \label{sec:method-semisupervised}

We follow the recent cross-modal pseudo-labeling approaches \cite{peng2021lidar, chong2022monodistill} to optimize the objective in \cref{eq:optimization}. For the training dataset, we optimize the student model with supervised learning using labels $y_T$. For unlabeled samples, predictions from the teacher model $f_t(x)$ are given as a proxy for $h_s(x)$, and the student model is optimized towards the proxy target. If we replace $h_s(x)$ with an optimal teacher model $h_t(x)$ in \cref{eq:optimization}, we can rewrite the unlabeled part as:
\begin{equation}\label{eq:ssl_lidar_aleatoric}
\sum_{(x_U,y_U) \epsilon {U}}{ (f_s(x_U;\theta)-h_t(x_U))^2 + u^{al}_{t}(x_U)}
\end{equation}

Using teacher model predictions as labels, the first term can theoretically be fully reduced after optimization. However, our initial objective of minimizing the expected loss of the student model does not converge toward its global minima due to the additional second term, aleatoric uncertainty of the teacher $u^{al}_{t}(x)$. With this formulation, we identify the potential error propagation from the teacher to the student model in our baseline framework \cite{peng2021lidar}. Therefore, in MonoLiG, we reduce the effect of pseudo-labels with high teacher aleatoric uncertainty while training the student model.

\cref{fig:overview} presents the training phase in MonoLiG with the following steps:
\begin{enumerate}
  \item Teacher training with aleatoric uncertainty using $T$
  \item Pseudo-label generation on samples of $U$ using the predictions of the teacher model
  \item Student training with $T$ using $y_T$ and with $U$ using pseudo-labels $f_t(x)$
\end{enumerate}

\textbf{Training of the teacher model with aleatoric uncertainty calculation.} We design MonoLiG to utilize any object detector as its teacher model. A typical 3D object detector outputs seven bounding box regression parameters defined by the center coordinates $(x,y,z)$, dimensions $(w,h,l)$, and rotation angle $\alpha$. We use gaussian modeling to model the aleatoric uncertainty of a bounding box. We assume a Gaussian distribution for each regression variable and modify the teacher detector to output the mean $\mu(x,W)$ and the uncertainty $\sigma^2(x,W)$. To optimize the teacher model with the uncertainty head, we use negative log-likelihood (NLL). For a Gaussian distribution, the NLL can be written as:
\begin{equation}
    L(x,W) = \frac{{y-\mu(x,W)}^2}{2\sigma^2(x,W)} + \frac{log{\sigma^2(x,W)}}{2}
\end{equation}

To define a single aleatoric uncertainty-based confidence score for a bounding box, we sum the uncertainty of the location regression parameters:
\begin{equation} \label{eq:aleatoric_of_box}
    u^{al}_{t}(x) = \sigma_x(x,W) + \sigma_y(x,W) + \sigma_z(x,W)
\end{equation}
\textbf{Pseudo-label and confidence generation.} To generate pseudo-labels for all samples in the unlabeled pool, we perform inference using our teacher model to detect objects and apply post-processing, such as non-maximum suppression (NMS). To scale the effect of teacher's predictions based on their uncertainty during the training of the student model, we assign a confidence score to each prediction. We propose to use $1-u^{al}_{t}(x)$ from \cref{eq:aleatoric_of_box} as the localization confidence and the probability of the predicted class $p(x)$ as the classification confidence and combine as follows:
\begin{equation}
    c(x) = (1 - u^{al}_{t}(x)) * p(x)
\end{equation}

\textbf{Student training using semi-supervised learning.} The MonoLiG framework is compatible with any object detector as its student model. The loss function is updated to incorporate the ability to scale the effect of each bounding box label based on its confidence. We scale the original loss of the student model with the confidence as follows:
\begin{equation}
    L_{c}(x,y,c) = c(x) * L(f_s(x),y)
\end{equation}

The student model is trained with labels $y_T$ for the training set $T$ and the pseudo-labels $\tilde{y}_U$ for the unlabeled set $U$. Joint loss is given as:
\begin{equation}
    L = L_{c}(x_T, y_T, c_T) + \lambda_U L_{c}(x_U, \tilde{y}_U, c_U)
\end{equation}
where $\lambda_U$ is the weight of the loss for unlabeled samples. In our experiments, we set $\lambda_U$ to $0.5$ and the confidence of labeled samples $c_T$ to $1$.

\subsection{Selection phase with active learning} \label{sec:method-activelearning}

The goal of the selection phase is to select the best subset $S^*$, such that after training with it results in a student model with a lower error than any other $S$ \cite{roy2001toward}.
\begin{equation}
    \forall S, E[L;\theta_{T+S^*}]<E[L;\theta_{T+S}]
\end{equation}

However, this requires re-training the model for every possible subset and evaluating the expectation, which is practically infeasible. LL4AL \cite{yoo2019learning} proposes a greedy solution by selecting samples with the highest loss from the unlabeled set $U$ and optimizing it with supervised learning after labeling. This way, the remaining set $U-S$ has a smaller expected loss. The greedy selection score $s(x)$ can be defined as,
\begin{equation}\label{eq:al_learningloss}
s(x) = L(f_s(x;\theta_T),y)
\end{equation}

Note that this criterion depends on the current model parameters $\theta_T$, trained with dataset $T$. Following Bayesian AL \cite{gal2017deep}, we argue that the optimal selection score should not depend on a specific parameter value but the expectation over the parameters $E_\theta$ for a weight distribution $p(\theta|D)$, due to the stochastic nature of training with random initialization and data shuffles. We decompose the loss-based criteria as follows (derivation in the supplement):
\begin{multline}\label{eq:al_c_as_loss}
s(x) = E_\theta[(f_s(x;\theta)-E_\theta[f_s(x;\theta)])^2]\\
+ (E_\theta[f_s(x;\theta)] - h_s(x))^2\\
+ u^{al}_{s}(x)\\
\end{multline}

Selecting based on the total loss, like LL4AL, leads to selecting samples with high aleatoric uncertainty $u^{al}_{s}$ that potentially harms the optimization. Therefore, we propose a selection criterion focusing on the first two components. The first component corresponds to the epistemic uncertainty of the student model. Using epistemic uncertainty as an AL scoring function is well-researched \cite{tsymbalov2018dropout, choi2021active}. However, previous AL methods cannot capture the second term without the ground-truth or $h_s(x)$. We propose using the teacher model's predictions as an estimate to $h_s(x)$ and define a new selection score, the inconsistency between the teacher and the student.

To have a coherent selection score with our semi-supervised training objective, we propose to select samples with high teacher aleatoric uncertainty $u^{al}_{t}$ for annotation instead of generating pseudo-labels. Recall that these samples harm the optimization of the student, and with this selection score, the remaining samples in $U$ contain pseudo-labels with high confidence.

\cref{fig:overview} presents the selection phase in MonoLiG consisting of epistemic uncertainty of the student model, teacher-student inconsistency, and aleatoric uncertainty of the teacher.

\textbf{Epistemic uncertainty of student model.} Following ensembling techniques to capture epistemic uncertainty \cite{beluch2018power, haussmann2020scalable}, we estimate $E_\theta$ using an ensemble of five models trained with different random initialization. Predictions from multiple ensemble members are matched using intersection over union (IoU), and the uncertainty of each box is represented by the total variance $u^{tv}_{s}(x)$ of its regression parameters.
\begin{equation}
u^{tv}_{s}(x) = E_\theta[(f_s(x;\theta)-E_\theta[f_s(x;\theta)])^2]
\end{equation}

\textbf{Teacher-student inconsistency.} Following the teacher-student paradigm in SSL, we propose to use $f_t(x)$ as an estimate to $h_s(x)$. Using the matching algorithm with IoU, we match predictions from the teacher with predictions from different ensemble members. Then we define teacher-student inconsistency $i_{ts}$ as the difference between the teacher model's regression parameters and the mean of the regression parameters from ensembles of student models:
\begin{equation}
i_{ts}(x) = (E_\theta[f_s(x;\theta)] - f_t(x))^2
\end{equation}

\textbf{Selection strategy.} We propose the selection score of MonoLiG as a combination of the three aforementioned scores as follows:
\begin{equation}\label{eq:selection_strategy}
s(x) = (u^{tv}_{s}(x) + i_{ts}(x)) * (u^{al}_{t}(x))
\end{equation}

We sum $u^{tv}_{s}$ and $i_{ts}(x)$ based on formulation in \cref{eq:al_c_as_loss} and multiply with $u^{al}_{t}$ due to difference in scales. Then, we aggregate the object-based score by taking the maximum score of objects to obtain a sample selection score.

\section{Experiments} \label{sec:experiments}

\subsection{Experimental setup} \label{sec:exp-setup}

\textbf{Datasets and evaluation metric.} We present our evaluation results on two autonomous driving datasets with synchronized LiDAR and camera frames and 3D bounding box labels: KITTI \cite{geiger2012we}, and the Waymo Open Dataset \cite{sun2020scalability}.

KITTI contains 7481 images for training and 7518 samples for testing. Since labels of the test set are unavailable, we further split the training set following \cite{lang2019pointpillars}, which results in 3712 training and 3769 validation samples. For the AL scoring comparison and the ablation study, we report on the validation set and present the performance of our semi-supervised training strategy on the test set. We report BEV AP and 3D AP with 40 recall points on the car class for moderate difficulty with a 0.7 IoU threshold. We also present results on a larger scale Waymo Open Dataset which contains 798 training and 202 validation sequences. Following CaDDN \cite{reading2021categorical}, we downsample the original training set by selecting every third frame, resulting in a training set of approximately 51K samples labeled with 3D bounding boxes. For the Waymo dataset, we present our results using the official Level 2 mAP metric with 0.5 IoU.

\textbf{Active learning details.} For KITTI, we randomly split the training set for each experiment into a 30\% labeled pool as an initial training dataset and a 70\% unlabeled pool. The initial 30\% of training data is used to pre-train the model. At each AL cycle, we compute scores on all samples in the unlabeled pool and select the 10\% with the highest score to add to the training set. To imitate labeling, we use the already available annotations. For Waymo, since it contains more samples, we start with a training set with 5\% samples and, at each cycle, add 5\%. We present the mean of the corresponding metrics for three experiments with three different random initial training datasets.

\textbf{Model architectures.} For our AL experiments, we use the SOTA DD3D \cite{park2021dd3d} as our student model and the well-established PV-RCNN \cite{shi2020pv} as our teacher model. We train for the same number of epochs and use the same hyperparameters and optimization scheme described in their respective papers. All experiments are conducted on an NVIDIA Tesla V100 GPU with PyTorch \cite{paszke2019pytorch}.

\subsection{Comparison with AL selection baselines}\label{sec:exp-ssal}

\begin{figure*}[t]
  \centering
  \begin{subfigure}{0.33\linewidth}
    \includegraphics[width=0.99\linewidth]{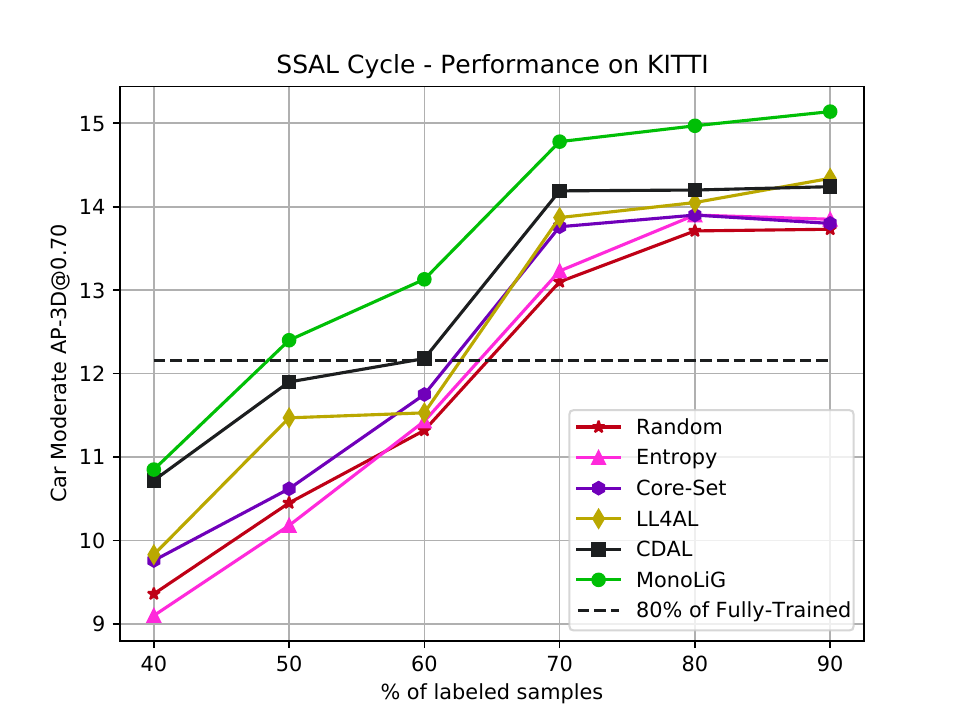}
    \caption{With semi-supervised training on KITTI}
    \label{fig:ssal_cycle-a}
  \end{subfigure}
  \hfill
  \begin{subfigure}{0.33\linewidth}
    \includegraphics[width=0.99\linewidth]{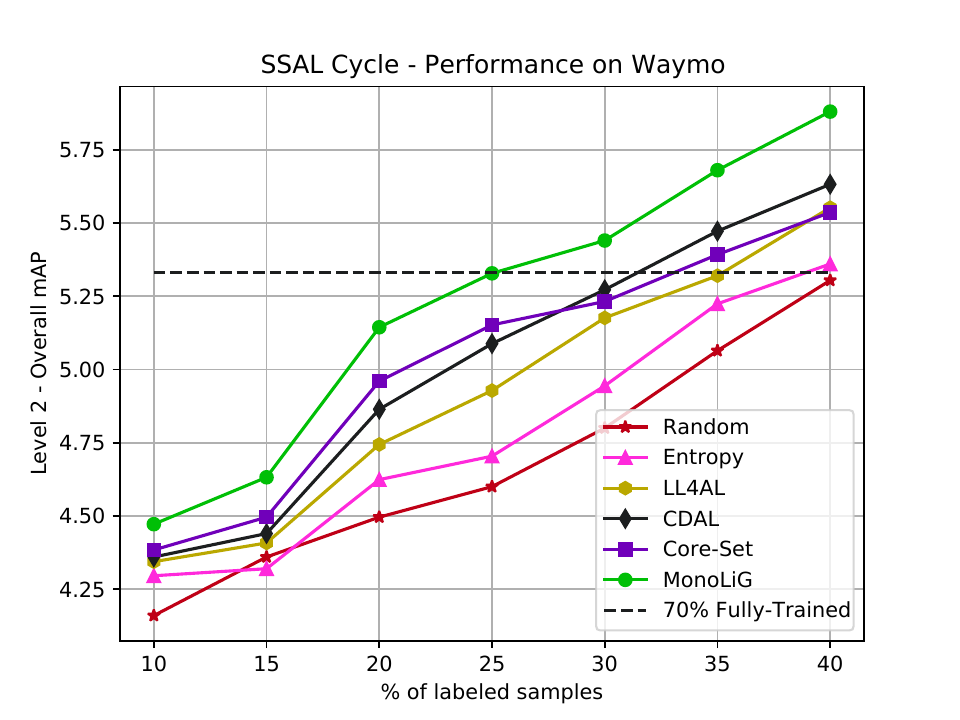}
    \caption{With semi-supervised training on Waymo}
    \label{fig:ssal_cycle-b}
  \end{subfigure}
  \hfill
  \begin{subfigure}{0.33\linewidth}
    \includegraphics[width=0.99\linewidth]{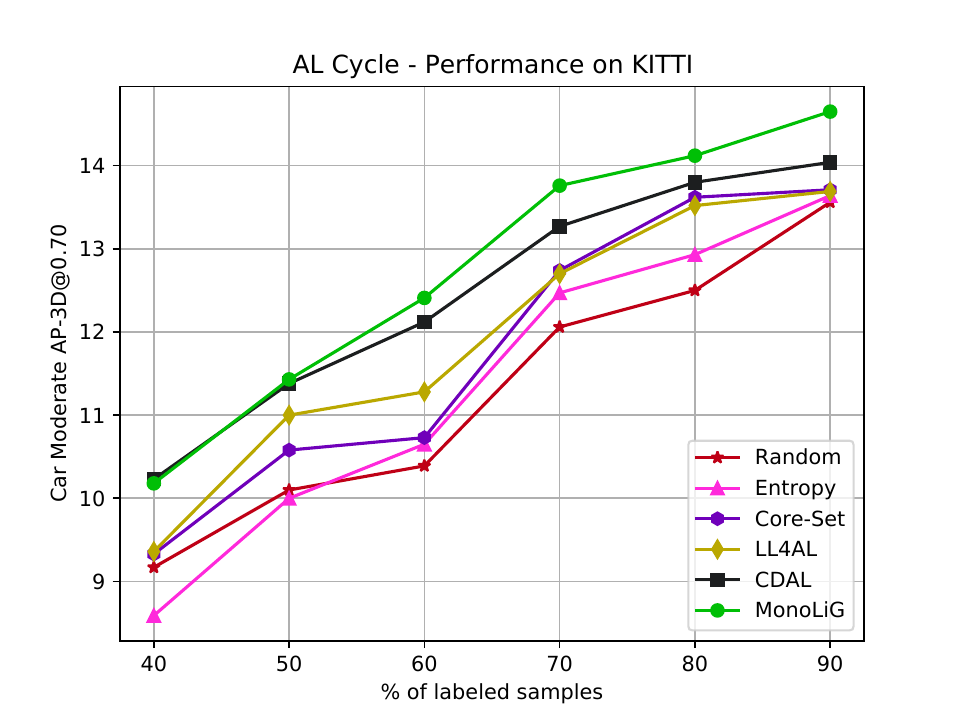}
    \caption{With supervised training on KITTI}
    \label{fig:ssal_cycle-c}
  \end{subfigure}
  \caption{Comparison with SOTA AL methods. Lines indicate the averaged results over three trials. Note that all methods start from the same network trained with the initial labeled data, corresponding to 30\% for KITTI and 5\% for Waymo.}
  \label{fig:ssal_cycle}
\end{figure*}

\textbf{With semi-supervised training.} To demonstrate the effectiveness of our MonoLiG framework, we perform an evaluation to other AL selection methods using the same semi-supervised training phase. Specifically, we compare our approach with six baseline methods: \textbf{Random} sampling, \textbf{Entropy} sampling, a diversity-based \textbf{Core-Set} method\cite{sener2018active}, a task-agnostic \textbf{LL4AL} method\cite{yoo2019learning}, and the state-of-the-art \textbf{CDAL} \cite{agarwal2020contextual} method. We also include the AP of a "fully-trained" detector, which is trained on the entire training set, to demonstrate the detector’s performance capability.

In \cref{fig:ssal_cycle-a}, we present the comparison with AL methods for KITTI. Our method outperforms all the uncertainty-based baselines by at least 1.02 3D AP in the first AL cycle. As the number of actively selected labels increases, our method outperforms Random by 1.61 and the second-best method, CDAL, by 0.75. In the final cycle, where we use 90\% of all the available data, with 60\% of it actively labeled, our method outperforms all methods by at least 6.32\%. Our approach reaches 80\% of the fully-trained performance using only 48\% of the data, compared to 60\% of CDAL and 65\% of random selection. This corresponds to a 17\% improvement in data savings. We consistently outperform LL4AL in all cycles, validating that our approach of decomposing the loss function and ignoring aleatoric uncertainty leads to a better selection strategy.

In \cref{fig:ssal_cycle-b}, we present the results for the Waymo dataset, which is larger and thus more intuitively benefits from AL data selection. Towards the end, we reach 4.4\% higher than the second-highest performing CDAL and 10.9\% higher than random selection. Our approach reaches 70\% of the fully-trained performance using only 25\% of available data, compared to CDAL at 32\% and random selection at 40\%, corresponding to an 15\% better data saving rate.

\textbf{With supervised training.} To evaluate the effectiveness of our selection criteria in the absence of a coherent training strategy, we conduct an evaluation by comparing it with the same baselines, but using only supervised learning. In \cref{fig:ssal_cycle-c}, our selection criterion achieves better results than other uncertainty-based selection scores by at least 0.77 3D AP. As we actively selected more samples, we reach 4.34\% higher than CDAL and 8.03\% higher than Random.

\subsection{Comparison with semi-supervised learning} \label{sec:exp-ssl}

\begin{figure*}[tb]
    \centering
    \includegraphics[width=0.9\linewidth]{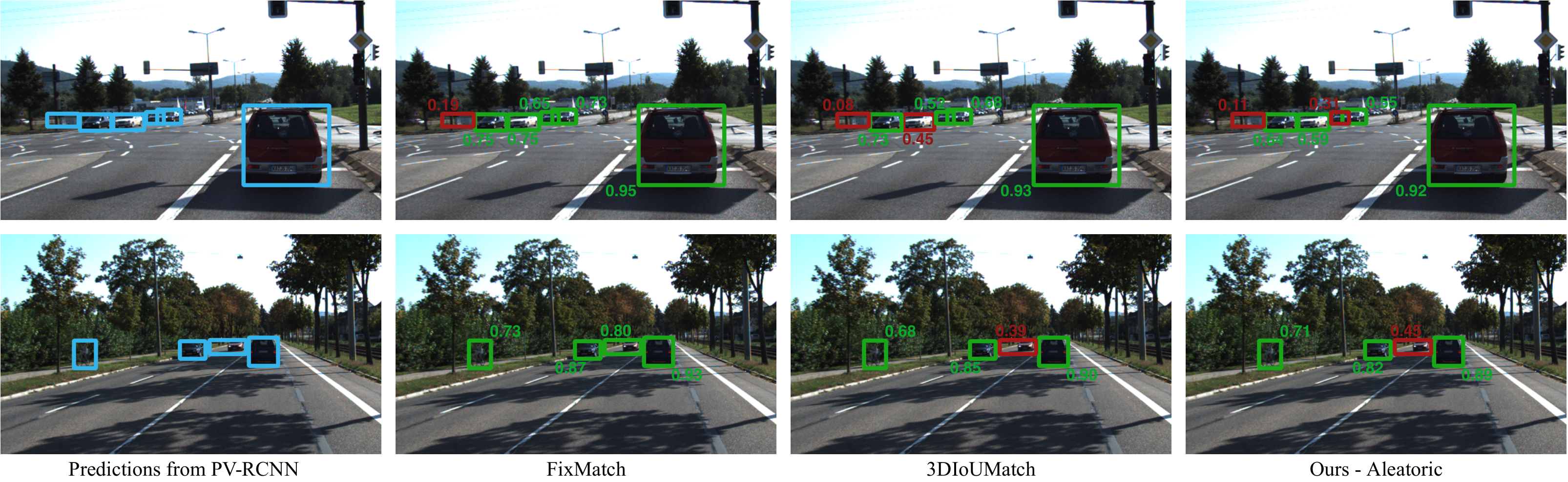}
    \caption{Qualitative comparisons of pseudo-label confidence algorithms (FixMatch, 3DIoUMatch, Ours) on KITTI. Blue boxes represent predictions from the PV-RCNN detector. Red and green boxes represent boxes with a confidence score below and above 0.5 threshold for the corresponding pseudo-label confidence approach.}
    \label{fig:filter_comparison}
\end{figure*}

We compare the performance of the monocular 3D detector on the KITTI \textit{test} to our semi-supervised training strategy. Following \cite{peng2021lidar}, we train with the entire KITTI training dataset and, as the unlabeled pool, use KITTI raw scenes, excluding the samples from the validation set. This dataset is called the KITTI-depth set and contains approximately 26K samples. For a fair comparison against LPCG-MonoFlex\cite{peng2021lidar}, we use MonoFlex as the student model.

\Cref{table:ssl_results} shows the results of our method compared to other SOTA monocular 3D detectors. Among the methods that use semi-supervised LiDAR guidance, our approach reaches +2.02 and +4.24 BEV AP than the SOTA LPCG and MonoDistill, respectively. Considering the performance weighted by the number of samples in each case, MonoLiG has a higher overall AP of 28.62 compared to 26.94 of LPCG. Using our semi-supervised training strategy, MonoFlex, proposed in 2019, lagging behind the current SOTA detector MonoDDE by 3.71 AP, reaches 3.37 higher performance.

\begin{table}[htbp]
    \centering
    \begin{tabular}{|l|c|ccc|}
    \hline
    Approaches & Extra & Mod. & Easy & Hard\\
    \hline\hline
    M3D-RPN \cite{brazil2019m3drpn} & - & 13.67 & 21.02 & 10.23  \\
    MonoRUn \cite{chen2021monorun} & - & 17.34 & 27.94 & 15.24  \\
    DDMP-3D \cite{wang2021depth} & KD & 17.89 & 28.08 & 13.44  \\
    PCT \cite{wang2021progressive} & KD & 19.03 & 29.65 & 15.92  \\
    MonoFlex \cite{zhen2021objects} & - & 19.75 & 28.23 & 16.89 \\
    MonoDTR \cite{huang2022monodtr} & - & 20.38 & 28.59 & 17.14 \\
    DID-M3D \cite{peng2022did}& - & 22.76 & 32.95 & 19.83 \\
    DD3D \cite{park2021dd3d} & DDAD & 23.41 & 32.35 & 20.42 \\
    MonoDDE \cite{ma2021delving} & - & 23.46 & 33.58 & 20.37 \\
    \hline
    MonoDistill \cite{chong2022monodistill} & - & 22.59 & 31.87 & 19.72 \\
    LPCG \cite{peng2021lidar} & KD & \underline{24.81} & \textbf{35.96} & \underline{21.86} \\
    \hline
    MonoLiG & KD & \textbf{26.83} & \underline{35.73} & \textbf{24.24} \\
    \hline
    \end{tabular}
    \caption{Comparison of BEV detection results on KITTI \textit{test} for monocular detectors. Note that both LPCG and MonoLiG use MonoFlex as the base detector. MonoDistill, LPCG, and MonoLiG are semi-supervised methods using additional information from LiDAR during training. KD and DDAD represent the extra datasets, KITTI-depth and DDAD15M\cite{park2021dd3d}, respectively.}
    \label{table:ssl_results}
\end{table}

We further investigate how our pseudo-label weighting strategy compares to other pseudo-label filtering strategies from the literature. We compare with \textbf{FixMatch} \cite{sohn2020fixmatch}, which uses the confidence score to filter out uncertain pseudo-label, and \textbf{3DIoUMatch} \cite{wang20213dioumatch}, which, in addition to the confidence score, uses an estimated IoU for filtering. We follow our approach of scaling the pseudo-labels with the corresponding uncertainty and present the results in \Cref{table:filtering_algos}. Our pseudo-label uncertainty approach reaches the highest performance, reaching 7.82 AP higher than the base model and 1.59 AP higher than 3DIoUMatch.
Furthermore, compared to 3DIoUMatch, our approach reduces additional time per iteration by 16\% and additional memory consumption by 3-fold. We observe that methods that consider localization uncertainty (3DIoUMatch, MonoLiG) perform better than methods that only use classification uncertainty (FixMatch).

In \cref{fig:filter_comparison}, we compare qualitatively how different pseudo-labeling uncertainty strategies work. In the first row, our approach and 3DIoUMatch filter out different vehicles, both difficult to localize for a monocular 3D object detector. We observe that our predictions correspond to more distant and occluded objects with low LiDAR point returns and, therefore, higher aleatoric uncertainty. Also, in the second row, the middle car with high bounding box localization error has low classification uncertainty but high localization uncertainty. For these types of objects, utilizing a strategy that also identifies localization uncertainty is essential.

\begin{table}[htbp]
    \centering
    \begin{tabular}{|l|c|c|c|}
    \hline
    Approaches & Mod. & Time (ms) & Memory\\
    \hline\hline
    DD3D \cite{park2021dd3d} & 16.92 & 53.4 & - \\
    \hline
    No Confidence & 20.14 & 53.4 & - \\
    FixMatch \cite{sohn2020fixmatch} & 22.59 & 53.4 & - \\
    3DIoUMatch \cite{wang20213dioumatch} & \underline{23.15} & 62.9(+17.8\%) & 696 MB \\
    Ours - Aleatoric & \textbf{24.74} & 54.2(+1.6\%) & 184 MB \\
    \hline
    \end{tabular}
    \caption{Comparison of different pseudo-labeling confidence algorithms on KITTI \textit{val}.}
    \label{table:filtering_algos}
\end{table}

\subsection{Ablation studies} \label{sec:exp-ablation}

\textbf{Ablation on scoring.} Next, we study the effect of each component in our MonoLiG framework. We present our ablation study with different combinations for our selection in the supervised-learning setting in \Cref{table:ablation-score}. Our findings indicate that, among the single scores, $i_{ts}$ performs best in the initial cycles, highlighting the effectiveness of utilizing teacher predictions when the student uncertainty is not yet well-learned. When all three scores are combined, we obtain the highest performing selection score.

\begin{table}[htbp]
    \centering
    \begin{tabular}{|l|cccc|}
    \hline
    $s(x)$ & 40\% & 50\% & 70\% & 90\% \\
    \hline\hline
    $u^{al}_{t}$ & 9.12 & 10.07 & 11.44 & 12.94 \\
    $u^{tv}_{s}$ & 8.86 & 10.51 & 12.64 & 14.12 \\
    $i_{ts}$ & \underline{9.53} & \underline{11.13} & 12.68 & 13.89 \\
    $u^{tv}_{s}+i_{ts}$ & 9.22 & 10.73 & \underline{13.39} & \underline{14.44} \\
    \hline
    $(u^{tv}_{s}+i_{ts})*u^{al}_{t}$ & \textbf{10.18} & \textbf{11.43} & \textbf{13.76} & \textbf{14.65} \\
    \hline
    \end{tabular}
    \caption{Ablation study on selection scores and training strategy on KITTI \textit{val}. The percentage indicates the ratio of labeled data.}
    \label{table:ablation-score}
\end{table}

\textbf{Different teacher-student architectures.} To show the robustness of MonoLiG to the architecture choice of the teacher and the student model, we try with two different monocular detectors: DD3D \cite{park2021dd3d} and MonoFlex \cite{zhen2021objects} as our student model and two different LiDAR detectors as our teacher model: PV-RCNN \cite{shi2020pv} and PointPillars \cite{lang2019pointpillars}. In \Cref{table:ablation-student}, we see that MonoLiG boosts the performance of both student models compared to the random sampling strategy. We observe a performance gain of 1.24 AP at 80\% data percentage for DD3D and even a further 1.85 AP for the MonoFlex. We also observe that for the choice of teacher model, the better the teacher model is, the higher the performance gain when MonoLiG is used.

\begin{table}[htbp]
    \centering
    \begin{tabular}{|l|cc|cc|}
    \hline
     & \multicolumn{2}{c|}{DD3D \cite{park2021dd3d}} & \multicolumn{2}{c|}{MonoFlex \cite{zhen2021objects}} \\
     & 40\% & 80\% & 40\% & 80\% \\
    \hline\hline
    Base & 7.58 & 12.47 & 5.22 & 10.08 \\
    \hline
    PointPillars \cite{lang2019pointpillars} & \underline{8.91} & \underline{12.89} & \underline{7.35} & \underline{10.92} \\
    PV-RCNN \cite{shi2020pv} & \textbf{9.36} & \textbf{13.71} & \textbf{7.81} & \textbf{11.33} \\
    \hline
    \end{tabular}
    \caption{Performance comparison of MonoLiG with two different LiDAR and monocular detectors on KITTI \textit{val}. The percentage indicates the ratio of labeled samples.}
    \label{table:ablation-student}
\end{table}

\section{Conclusion} \label{sec:conclusion}

We introduced a novel SSAL framework. MonoLiG consists of a novel training phase that uses aleatoric uncertainty weighted pseudo-labels from the LiDAR detector to guide the training of the monocular detector and a selection phase with a novel acquisition function based on the inconsistency between predictions from the LiDAR and the monocular detector. Our extensive experiments validate the effectiveness of MonoLiG compared to both AL and SSL baselines. We further showed that MonoLiG could easily be adapted to any monocular detector. Pseudo-labels' quality is essential for our framework; we will further explore how to generate more precise pseudo-labels by adding more modalities, e.g., radar and tracking over time.

{\small
\bibliographystyle{ieee_fullname}
\bibliography{references}

\begin{thebibliography}{10}\itemsep=-1pt

\bibitem{agarwal2020contextual}
Sharat Agarwal, Himanshu Arora, Saket Anand, and Chetan Arora.
\newblock Contextual diversity for active learning.
\newblock In {\em ECCV}, 2020.

\bibitem{aghdam2019active}
Hamed~H Aghdam, Abel Gonzalez-Garcia, Joost van~de Weijer, and Antonio~M
  L{\'o}pez.
\newblock Active learning for deep detection neural networks.
\newblock In {\em ICCV}, 2019.

\bibitem{arazo2020pseudo}
Eric Arazo, Diego Ortego, Paul Albert, Noel~E O’Connor, and Kevin McGuinness.
\newblock Pseudo-labeling and confirmation bias in deep semi-supervised
  learning.
\newblock In {\em IJCNN}, 2020.

\bibitem{beluch2018power}
William~H Beluch, Tim Genewein, Andreas N{\"u}rnberger, and Jan~M K{\"o}hler.
\newblock The power of ensembles for active learning in image classification.
\newblock In {\em CVPR}, 2018.

\bibitem{brazil2019m3drpn}
Garrick Brazil and Xiaoming Liu.
\newblock M3d-rpn: Monocular 3d region proposal network for object detection.
\newblock In {\em ICCV}, 2019.

\bibitem{caine2021pseudo}
Benjamin Caine, Rebecca Roelofs, Vijay Vasudevan, Jiquan Ngiam, Yuning Chai,
  Zhifeng Chen, and Jonathon Shlens.
\newblock Pseudo-labeling for scalable 3d object detection.
\newblock {\em arXiv preprint arXiv:2103.02093}, 2021.

\bibitem{chen2020temporal}
Cong Chen, Shouyang Dong, Ye Tian, Kunlin Cao, Li Liu, and Yuanhao Guo.
\newblock Temporal self-ensembling teacher for semi-supervised object
  detection.
\newblock {\em Transactions on Multimedia}, 2021.

\bibitem{chen2021monorun}
Hansheng Chen, Yuyao Huang, Wei Tian, Zhong Gao, and Lu Xiong.
\newblock Monorun: Monocular 3d object detection by reconstruction and
  uncertainty propagation.
\newblock In {\em CVPR}, 2021.

\bibitem{choi2021active}
Jiwoong Choi, Ismail Elezi, Hyuk-Jae Lee, Cl{\'e}ment Farabet, and
  Jos{\'e}~Manuel {\'A}lvarez.
\newblock Active learning for deep object detection via probabilistic modeling.
\newblock In {\em ICCV}, 2021.

\bibitem{chong2022monodistill}
Zhiyu Chong, Xinzhu Ma, Hong Zhang, Yuxin Yue, Haojie Li, Zhihui Wang, and
  Wanli Ouyang.
\newblock Monodistill: Learning spatial features for monocular 3d object
  detection.
\newblock In {\em ICLR}, 2022.

\bibitem{elezi2022not}
Ismail Elezi, Zhiding Yu, Anima Anandkumar, Laura Leal-Taixe, and Jose~M
  Alvarez.
\newblock Not all labels are equal: Rationalizing the labeling costs for
  training object detection.
\newblock In {\em CVPR}, 2022.

\bibitem{feng2021review}
Di Feng, Ali Harakeh, Steven~L Waslander, and Klaus Dietmayer.
\newblock A review and comparative study on probabilistic object detection in
  autonomous driving.
\newblock {\em T-ITS}, 2021.

\bibitem{feng2019deep}
Di Feng, Xiao Wei, Lars Rosenbaum, Atsuto Maki, and Klaus Dietmayer.
\newblock Deep active learning for efficient training of a lidar 3d object
  detector.
\newblock In {\em IV}, 2019.

\bibitem{gal2017deep}
Yarin Gal, Riashat Islam, and Zoubin Ghahramani.
\newblock Deep bayesian active learning with image data.
\newblock In {\em ICML}, 2017.

\bibitem{gao2020consistency}
Mingfei Gao, Zizhao Zhang, Guo Yu, Sercan~{\"O} Ar{\i}k, Larry~S Davis, and
  Tomas Pfister.
\newblock Consistency-based semi-supervised active learning: Towards minimizing
  labeling cost.
\newblock In {\em ECCV}, 2020.

\bibitem{geiger2012we}
Andreas Geiger, Philip Lenz, and Raquel Urtasun.
\newblock Are we ready for autonomous driving? the kitti vision benchmark
  suite.
\newblock In {\em CVPR}, 2012.

\bibitem{haussmann2020scalable}
Elmar Haussmann, Michele Fenzi, Kashyap Chitta, Jan Ivanecky, Hanson Xu, Donna
  Roy, Akshita Mittel, Nicolas Koumchatzky, Clement Farabet, and Jose~M
  Alvarez.
\newblock Scalable active learning for object detection.
\newblock In {\em IV}, 2020.

\bibitem{houlsby2011bayesian}
Neil Houlsby, Ferenc Husz{\'a}r, Zoubin Ghahramani, and M{\'a}t{\'e} Lengyel.
\newblock Bayesian active learning for classification and preference learning.
\newblock {\em arXiv preprint arXiv:1112.5745}, 2011.

\bibitem{huang2022monodtr}
Kuan-Chih Huang, Tsung-Han Wu, Hung-Ting Su, and Winston~H. Hsu.
\newblock Monodtr: Monocular 3d object detection with depth-aware transformer.
\newblock In {\em CVPR}, 2022.

\bibitem{huang2021semi}
Siyu Huang, Tianyang Wang, Haoyi Xiong, Jun Huan, and Dejing Dou.
\newblock Semi-supervised active learning with temporal output discrepancy.
\newblock In {\em ICCV}, 2021.

\bibitem{jeong2019consistency}
Jisoo Jeong, Seungeui Lee, Jeesoo Kim, and Nojun Kwak.
\newblock Consistency-based semi-supervised learning for object detection.
\newblock In {\em NeurIPS}, 2019.

\bibitem{laine2017temporal}
Samuli Laine and Timo Aila.
\newblock Temporal ensembling for semi-supervised learning.
\newblock In {\em ICLR}, 2017.

\bibitem{lang2019pointpillars}
Alex~H Lang, Sourabh Vora, Holger Caesar, Lubing Zhou, Jiong Yang, and Oscar
  Beijbom.
\newblock Pointpillars: Fast encoders for object detection from point clouds.
\newblock In {\em CVPR}, 2019.

\bibitem{lee2013pseudo}
Dong-Hyun Lee et~al.
\newblock Pseudo-label: The simple and efficient semi-supervised learning
  method for deep neural networks.
\newblock In {\em ICMLW}, 2013.

\bibitem{ma2021delving}
Xinzhu Ma, Yinmin Zhang, Dan Xu, Dongzhan Zhou, Shuai Yi, Haojie Li, and Wanli
  Ouyang.
\newblock Delving into localization errors for monocular 3d object detection.
\newblock In {\em CVPR}, 2021.

\bibitem{mugnai2021soft}
Daniele Mugnai, Federico Pernici, Francesco Turchini, and Alberto~Del Bimbo.
\newblock Soft pseudo-labeling semi-supervised learning applied to fine-grained
  visual classification.
\newblock In {\em ICPR}, 2021.

\bibitem{park2021dd3d}
Dennis Park, Rares Ambrus, Vitor Guizilini, Jie Li, and Adrien Gaidon.
\newblock Is pseudo-lidar needed for monocular 3d object detection?
\newblock In {\em ICCV}, 2021.

\bibitem{paszke2019pytorch}
Adam Paszke, Sam Gross, Francisco Massa, Adam Lerer, James Bradbury, Gregory
  Chanan, Trevor Killeen, Zeming Lin, Natalia Gimelshein, Luca Antiga, et~al.
\newblock Pytorch: An imperative style, high-performance deep learning library.
\newblock In {\em NeurIPS}, 2019.

\bibitem{peng2021lidar}
Liang Peng, Fei Liu, Zhengxu Yu, Senbo Yan, Dan Deng, Zheng Yang, Haifeng Liu,
  and Deng Cai.
\newblock Lidar point cloud guided monocular 3d object detection.
\newblock In {\em ECCV}, 2022.

\bibitem{peng2022did}
Liang Peng, Xiaopei Wu, Zheng Yang, Haifeng Liu, and Deng Cai.
\newblock Did-m3d: Decoupling instance depth for monocular 3d object detection.
\newblock In {\em ECCV}, 2022.

\bibitem{reading2021categorical}
Cody Reading, Ali Harakeh, Julia Chae, and Steven~L Waslander.
\newblock Categorical depth distribution network for monocular 3d object
  detection.
\newblock In {\em CVPR}, 2021.

\bibitem{rizve2021defense}
Mamshad~Nayeem Rizve, Kevin Duarte, Yogesh~S Rawat, and Mubarak Shah.
\newblock In defense of pseudo-labeling: An uncertainty-aware pseudo-label
  selection framework for semi-supervised learning.
\newblock In {\em ICLR}, 2021.

\bibitem{roy2001toward}
Nicholas Roy and Andrew McCallum.
\newblock Toward optimal active learning through monte carlo estimation of
  error reduction.
\newblock In {\em ICML}, 2001.

\bibitem{schmidt2020advanced}
Sebastian Schmidt, Qing Rao, Julian Tatsch, and Alois Knoll.
\newblock Advanced active learning strategies for object detection.
\newblock In {\em IV}, 2020.

\bibitem{sener2018active}
Ozan Sener and Silvio Savarese.
\newblock Active learning for convolutional neural networks: A core-set
  approach.
\newblock In {\em ICLR}, 2018.

\bibitem{shannon1948mathematical}
Claude~Elwood Shannon.
\newblock A mathematical theory of communication.
\newblock {\em Mobile Computing and Communications Review}, 2001.

\bibitem{shi2020pv}
Shaoshuai Shi, Chaoxu Guo, Li Jiang, Zhe Wang, Jianping Shi, Xiaogang Wang, and
  Hongsheng Li.
\newblock Pv-rcnn: Point-voxel feature set abstraction for 3d object detection.
\newblock In {\em CVPR}, 2020.

\bibitem{shi2018transductive}
Weiwei Shi, Yihong Gong, Chris Ding, Zhiheng~MaXiaoyu Tao, and Nanning Zheng.
\newblock Transductive semi-supervised deep learning using min-max features.
\newblock In {\em ECCV}, 2018.

\bibitem{simeoni2021rethinking}
Oriane Sim{\'e}oni, Mateusz Budnik, Yannis Avrithis, and Guillaume Gravier.
\newblock Rethinking deep active learning: Using unlabeled data at model
  training.
\newblock In {\em ICPR}, 2021.

\bibitem{sinha2019variational}
Samarth Sinha, Sayna Ebrahimi, and Trevor Darrell.
\newblock Variational adversarial active learning.
\newblock In {\em ICCV}, 2019.

\bibitem{sohn2020fixmatch}
Kihyuk Sohn, David Berthelot, Nicholas Carlini, Zizhao Zhang, Han Zhang,
  Colin~A Raffel, Ekin~Dogus Cubuk, Alexey Kurakin, and Chun-Liang Li.
\newblock Fixmatch: Simplifying semi-supervised learning with consistency and
  confidence.
\newblock In {\em NeurIPS}, 2020.

\bibitem{song2019combining}
Shuang Song, David Berthelot, and Afshin Rostamizadeh.
\newblock Combining mixmatch and active learning for better accuracy with fewer
  labels.
\newblock {\em arXiv preprint arXiv:1912.00594}, 2019.

\bibitem{sun2020scalability}
Pei Sun, Henrik Kretzschmar, Xerxes Dotiwalla, Aurelien Chouard, Vijaysai
  Patnaik, Paul Tsui, James Guo, Yin Zhou, Yuning Chai, Benjamin Caine, Vijay
  Vasudevan, Wei Han, Jiquan Ngiam, Hang Zhao, Aleksei Timofeev, Scott
  Ettinger, Maxim Krivokon, Amy Gao, Aditya Joshi, Yu Zhang, Jonathon Shlens,
  Zhifeng Chen, and Dragomir Anguelov.
\newblock Scalability in perception for autonomous driving: Waymo open dataset.
\newblock In {\em CVPR}, 2020.

\bibitem{tarvainen2017mean}
Antti Tarvainen and Harri Valpola.
\newblock Mean teachers are better role models: Weight-averaged consistency
  targets improve semi-supervised deep learning results.
\newblock In {\em NeurIPS}, 2017.

\bibitem{tsymbalov2018dropout}
Evgenii Tsymbalov, Maxim Panov, and Alexander Shapeev.
\newblock Dropout-based active learning for regression.
\newblock In {\em AIST}, 2018.

\bibitem{wang20213dioumatch}
He Wang, Yezhen Cong, Or Litany, Yue Gao, and Leonidas~J Guibas.
\newblock 3dioumatch: Leveraging iou prediction for semi-supervised 3d object
  detection.
\newblock In {\em CVPR}, 2021.

\bibitem{wang2020semi}
Jun Wang, Shaoguo Wen, Kaixing Chen, Jianghua Yu, Xin Zhou, Peng Gao,
  Changsheng Li, and Guotong Xie.
\newblock Semi-supervised active learning for instance segmentation via scoring
  predictions.
\newblock In {\em BMVC}, 2020.

\bibitem{wang2021depth}
Li Wang, Liang Du, Xiaoqing Ye, Yanwei Fu, Guodong Guo, Xiangyang Xue, Jianfeng
  Feng, and Li Zhang.
\newblock Depth-conditioned dynamic message propagation for monocular 3d object
  detection.
\newblock In {\em CVPR}, 2021.

\bibitem{wang2021progressive}
Li Wang, Li Zhang, Yi Zhu, Zhi Zhang, Tong He, Mu Li, and Xiangyang Xue.
\newblock Progressive coordinate transforms for monocular 3d object detection.
\newblock In {\em NeurIPS}, 2021.

\bibitem{yoo2019learning}
Donggeun Yoo and In~So Kweon.
\newblock Learning loss for active learning.
\newblock In {\em CVPR}, 2019.

\bibitem{yu2022consistency}
Weiping Yu, Sijie Zhu, Taojiannan Yang, and Chen Chen.
\newblock Consistency-based active learning for object detection.
\newblock In {\em CVPR}, 2022.

\bibitem{yuan2021multiple}
Tianning Yuan, Fang Wan, Mengying Fu, Jianzhuang Liu, Songcen Xu, Xiangyang Ji,
  and Qixiang Ye.
\newblock Multiple instance active learning for object detection.
\newblock In {\em CVPR}, 2021.

\bibitem{zhen2021objects}
Yunpeng Zhang, Jiwen Lu, and Jie Zhou.
\newblock Objects are different: Flexible monocular 3d object detection.
\newblock In {\em CVPR}, 2021.

\end{thebibliography}
}

\clearpage

\twocolumn[{
\renewcommand\twocolumn[1][]{#1}
\centering
\Large
\textbf{Monocular 3D Object Detection with LiDAR Guided Semi Supervised Active Learning} \\
\vspace{0.5em}Supplementary Material \\
\vspace{1.0em}

}]

\setcounter{section}{0}

\section{Derivation of \cref{eqn:loss+alea}}\label{sec:Eq2}

For simplicity, we use $f$ as $f_s$ and rewrite \cref{eqn:optimization_integral} with the MSE loss function as follows:

\begin{equation}\label{eq:integral-mse}
E[L] = \iint{(f(x)-y)^2 dxdy}
\end{equation}

The optimization goal is to find an f(x) that minimizes $E[L]$. If we assume a completely flexible function f(x), we can do this formally by taking the derivative to give

\begin{equation}
\frac{\delta E[L]}{\delta f(x)} = 2 \int{(f(x)-y) dy} = 0
\end{equation}

We define $h(x)$ as the optimal function that satisfies this equation. Adding and subtracting $h(x)$ to \cref{eq:integral-mse} gives

\begin{equation}
E[L] = \iint{(f(x)-h(x)+h(x)-y)^2 dxdy}
\end{equation}
\begin{multline}
E[L] = \iint{(f(x)-h(x))^2 + (h(x)-y)^2 dxdy}\\
+ 2 \iint{(f(x)-h(x)) (h(x)-y) dxdy}
\end{multline}

where the final term is zero because for the optimal network $\int{(h(x)-y) dy} = 0$. Therefore, for a single point $x$, we can write the loss function as \cref{eqn:loss+alea}.

\section{Derivation of \cref{eq:al_c_as_loss}}\label{sec:Eq12}
We start from the loss definition in \cref{eqn:loss+alea}. For simplicity of notation, we use $f(x)$ instead of $f_s(x;\theta)$ and define expectation over the model parameters $E_{\theta}[f_s(x;\theta)]$ as $\mu$. Adding and subtracting $\mu$ to \cref{eqn:loss+alea} and expanding results in the following equation:
\begin{multline}
L(x) = (f(x)-\mu)^2 + (\mu-h(x))^2\\
+ 2 * (f(x)-\mu) (\mu-h(x))\\
+ (h(x)-y)^2\\
\end{multline}

For reasons mentioned in the paper, we take the expectation of the loss over the model parameters $E_\theta$.
\begin{multline}\label{eq:exp}
E_\theta[L(x)] = E_\theta[(f(x)-\mu)^2] + (\mu - h(x))^2\\
+ 2 * E_\theta[ (f(x)-\mu) (\mu-h(x))]\\
+ (h(x)-y)^2\\
\end{multline}

Note that the third term disappears, as shown below:
\begin{multline}
E_\theta[ (f(x)-\mu) (\mu-h(x))] \\
= E_\theta[ f(x)*\mu - h(x)*f(x) - \mu^2 + \mu*h(x)]\\
= \mu^2 - h(x)*\mu - \mu^2 + \mu*h(x) = 0
\end{multline}

Then \cref{eq:exp} is written as,
\begin{multline}\label{eq:as-f}
E_\theta[L(x)] = E_\theta[(f(x)-\mu)^2] + (\mu - h(x))^2 + (h(x)-y)^2
\end{multline}

where $(h(x)-y)^2$ is the aleatoric uncertainty $u^{al}$. For the student model $f_s(x,\theta)$, \cref{eq:as-f} results in \cref{eq:al_c_as_loss}.

\section{Statistics of \cref{eq:selection_strategy})}\label{sec:statistics_eq_15}

We provide the statistical summary for the components in \cref{eq:selection_strategy} from the main paper. $u^{tv}_{s}$ and $i_{ts}$ have a range from 0 to 3.49, a mean of 0.87, and a variance of 0.28. Similarly, $u^{al}_{t}$ has a range from 0.19 to 1.23, a mean of 0.25, and a variance of 0.16.

\section{Hyperparameters}
\subsection{Weight of Unlabeled Samples $\lambda_U$}

\begin{table}[ht]
    \centering
    \begin{tabular}{|l|ccc|}
    \hline
    $\lambda_U$ & Mod. & Easy & Hard\\
    \hline\hline
    $0.1$ & 24.48 & 34.51 & 21.37 \\
    $0.2$ & 26.83 & \underline{36.33} & 24.04 \\
    $0.5$ & \textbf{27.92} & \textbf{36.86} & \underline{26.03} \\
    $0.7$ & \underline{27.37} & 35.28 & \textbf{26.51} \\
    $1.0$ & 25.77 & 35.02 & 24.69 \\
    $2.0$ & 24.38 & 34.94 & 22.34 \\
    \hline
    \end{tabular}
    \caption{The effect of hyperparameter $\lambda_u$ on the BEV AP performance of the monocular detector on KITTI \textit{val}}
    \label{table:supp-weight}
\end{table}

\cref{table:supp-weight} reports the experimental results on finetuning the hyperparameter for the unsupervised loss weight $\lambda_U$. We discovered that having a small or large $\lambda_U$ drops the performance. Therefore, finding a balanced ratio between supervised and unsupervised loss is important. Based on our findings in the ablation study, we set $\lambda_U = 0.5$ for all trainings in the main paper.

\section{Effect of Aleatoric Head on LiDAR Detector}\label{sec:aleatoric_head}

We investigate the impact of aleatoric uncertainty on the performance of the PV-RCNN LiDAR detector, as presented in \cref{table:supp-aleatoric}. Our findings indicate that aleatoric uncertainty has a negligible effect on the detector's performance, as evidenced by the slight decrease of only 0.15 and 0.25 in the moderate and easy AP scores, respectively, and a modest increase of 0.09 in the hard AP score.

\begin{table}[htbp]
    \centering
    \begin{tabular}{|c|ccc|}
    \hline
    Aleatoric & Mod. & Easy & Hard \\
    \hline\hline
    X & 82.58 & 89.95 & 77.32 \\
    \checkmark & 82.43 & 89.70 & 77.41 \\
    \hline
    \end{tabular}
    \caption{Comparison of aleatoric uncertainty head on the BEV AP performance of the LiDAR detector on KITTI \textit{val}.}
    \label{table:supp-aleatoric}
\end{table}

\section{Effect of Extra Data on KITTI Classes}

We present the effect of the number of extra samples used for semi-supervised learning on the performance of the MonoFlex detector \cite{zhen2021objects} in \cref{table:supp-kitti-extended}. We observe that as more data is trained with our semi-supervised strategy, performance for the \textit{Car} and \textit{Cyclist} classes increases. \textit{Pedestrian} performance is affected less by the semi-supervised training and even decreases for certain experiments, and we attribute this to LiDAR having a lower performance on the \textit{Pedestrian} class compared to the other two classes. This solidifies our conclusion in the main paper that as the performance of the teacher model increases, we expect better performance on the student model.

\begin{table*}[htbp]
    \centering
    \begin{tabular}{|l|ccc|ccc|ccc|}
    \hline
    \# of Extra& \multicolumn{3}{c|}{Vehicle} & \multicolumn{3}{c|}{Pedestrian} & \multicolumn{3}{c|}{Cyclist}\\
     & Mod. & Easy & Hard & Mod & Easy & Hard & Mod. & Easy & Hard\\
    \hline\hline
    0\%  & 14.96 & 20.09 & 13.62 & 6.39 & 8.76 & 4.75 & 3.30 & 4.49 & 2.58 \\
    10\% & 15.24 & 20.46 & 13.72 & 6.24 & 8.43 & 4.41 & 3.14 & 4.38 & 2.44 \\
    20\% & 15.82 & 20.41 & 14.42 & \textbf{6.42} & 8.34 & \textbf{4.87} & 3.32 & 4.67 & 2.50 \\
    30\% & 15.63 & 20.75 & 14.46 & 6.36 & \textbf{8.97} & 4.26 & 3.60 & 5.10 & 2.89 \\
    40\% & 16.55 & 21.39 & 14.94 & 6.02 & 8.79 & 4.12 & 3.85 & 5.20 & 2.74 \\
    50\% & \textbf{16.90} & \textbf{22.07} & \textbf{15.27} & 6.25 & 8.88 & 4.38 & \textbf{4.10} & \textbf{5.64} & \textbf{2.96} \\
    \hline
    \end{tabular}
    \caption{Effect of number of unlabeled samples in the 3D AP performance. We use KITTI-depth dataset explained in the main paper and randomly select a portion of it for each experiment.}
    \label{table:supp-kitti-extended}
\end{table*}

\section{Exact Values from AL Figures}

Due to the limited space in the main paper, we present our AL comparisons as plots. \cref{table:supp-kitti-fig4a}, \cref{table:supp-kitti-fig4b}, \cref{table:supp-kitti-fig4c} provides the exact metric values for Figures \ref{table:supp-kitti-fig4a}, \ref{table:supp-kitti-fig4b}, \ref{table:supp-kitti-fig4c} from the main paper. The mean and variances of three experiments trained with different random initializations are presented.

\begin{table*}[htbp]
    \centering
    \begin{tabular}{|l|ccccccc|}
    \hline
     & 30 & 40 & 50 & 60 & 70 & 80 & 90 \\
    \hline\hline
    Random & 8.04$\pm$0.23 & 9.36$\pm$0.16 & 10.45$\pm$0.23 & 11.32$\pm$0.15 & 13.10$\pm$0.10 & 13.71$\pm$0.13 & 13.73$\pm$0.25 \\
    Entropy & 8.04$\pm$0.23 & 9.10$\pm$0.30 & 10.18$\pm$0.10 & 11.43$\pm$0.25 & 13.23$\pm$0.18 & 13.90$\pm$0.12 & 13.85$\pm$0.25 \\
    Core-Set & 8.04$\pm$0.23 & 9.76$\pm$0.16 & 10.62$\pm$0.10 & 11.75$\pm$0.16 & 13.76$\pm$0.27 & 13.90$\pm$0.10 & 13.80$\pm$0.25 \\
    LL4AL & 8.04$\pm$0.23 & 9.83$\pm$0.22 & 11.47$\pm$0.11 & 11.53$\pm$0.13 & 13.87$\pm$0.15 & 14.05$\pm$0.27 & 14.34$\pm$0.17 \\
    CDAL & 8.04$\pm$0.23 & 10.72$\pm$0.24 & 11.90$\pm$0.17 & 12.18$\pm$0.13 & 14.19$\pm$0.26 & 14.20$\pm$0.19 & 14.24$\pm$0.19 \\
    MonoLiG & 8.04$\pm$0.23 & 10.85$\pm$0.17 & 12.40$\pm$0.16 & 13.13$\pm$0.19 & 14.78$\pm$0.11 & 14.97$\pm$0.13 & 15.14$\pm$0.20 \\
    \hline
    \end{tabular}
    \caption{Comparison with SOTA AL methods with semi-supervised training on KITTI \textit{val}.}
    \label{table:supp-kitti-fig4a}
\end{table*}

\begin{table*}[htbp]
    \centering
    \begin{tabular}{|l|cccccccc|}
    \hline
     & 5 & 10 & 15 & 20 & 25 & 30 & 35 & 40 \\
    \hline\hline
    Random & 3.50$\pm$0.10 & 4.16$\pm$0.07 & 4.36$\pm$0.13 & 4.50$\pm$0.05 & 4.60$\pm$0.13 & 4.80$\pm$0.07 & 5.06$\pm$0.06 & 5.30$\pm$0.08 \\
    Entropy & 3.50$\pm$0.10 & 4.30$\pm$0.08 & 4.32$\pm$0.12 & 4.62$\pm$0.12 & 4.70$\pm$0.12 & 4.94$\pm$0.11 & 5.22$\pm$0.12 & 5.36$\pm$0.07 \\
    LL4AL & 3.50$\pm$0.10 & 4.34$\pm$0.14 & 4.41$\pm$0.08 & 4.74$\pm$0.07 & 4.93$\pm$0.10 & 5.18$\pm$0.12 & 5.32$\pm$0.10 & 5.55$\pm$0.15 \\
    CDAL & 3.50$\pm$0.10 & 4.36$\pm$0.13 & 4.44$\pm$0.14 & 4.86$\pm$0.11 & 5.09$\pm$0.10 & 5.27$\pm$0.06 & 5.47$\pm$0.10 & 5.63$\pm$0.14 \\
    Core-Set & 3.50$\pm$0.10 & 4.38$\pm$0.15 & 4.50$\pm$0.15 & 4.96$\pm$0.06 & 5.15$\pm$0.11 & 5.23$\pm$0.07 & 5.39$\pm$0.11 & 5.54$\pm$0.14 \\
    MonoLiG & 3.50$\pm$0.10 & 4.47$\pm$0.05 & 4.63$\pm$0.11 & 5.14$\pm$0.14 & 5.33$\pm$0.07 & 5.44$\pm$0.11 & 5.68$\pm$0.06 & 5.88$\pm$0.09 \\
    \hline
    \end{tabular}
    \caption{Comparison with SOTA AL methods with semi-supervised training on Waymo \textit{val}.}
    \label{table:supp-kitti-fig4b}
\end{table*}

\begin{table*}[htbp]
    \centering
    \begin{tabular}{|l|ccccccc|}
    \hline
     & 30 & 40 & 50 & 60 & 70 & 80 & 90 \\
    \hline\hline
    Random & 7.81$\pm$0.18 & 9.17$\pm$0.26 & 10.10$\pm$0.21 & 10.39$\pm$0.27 & 12.06$\pm$0.21 & 12.50$\pm$0.27 & 13.56$\pm$0.16 \\
    Entropy & 7.81$\pm$0.18 & 8.59$\pm$0.19 & 10.00$\pm$0.17 & 10.65$\pm$0.30 & 12.47$\pm$0.21 & 12.93$\pm$0.15 & 13.64$\pm$0.11 \\
    Core-Set & 7.81$\pm$0.18 & 9.33$\pm$0.11 & 10.58$\pm$0.21 & 10.73$\pm$0.25 & 12.74$\pm$0.20 & 13.62$\pm$0.14 & 13.71$\pm$0.21 \\
    LL4AL & 7.81$\pm$0.18 & 9.36$\pm$0.21 & 11.00$\pm$0.26 & 11.28$\pm$0.15 & 12.70$\pm$0.13 & 13.52$\pm$0.26 & 13.69$\pm$0.10 \\
    CDAL & 7.81$\pm$0.18 & 10.23$\pm$0.26 & 11.38$\pm$0.13 & 12.12$\pm$0.23 & 13.27$\pm$0.12 & 13.80$\pm$0.17 & 14.04$\pm$0.22 \\
    MonoLiG & 7.81$\pm$0.18 & 10.18$\pm$0.22 & 11.43$\pm$0.19 & 12.41$\pm$0.18 & 13.76$\pm$0.21 & 14.12$\pm$0.14 & 14.65$\pm$0.10 \\
    \hline
    \end{tabular}
    \caption{Comparison with SOTA AL methods with supervised training on KITTI \textit{val}.}
    \label{table:supp-kitti-fig4c}
\end{table*}

\section{Qualitative Results on KITTI}

\begin{figure*}
  \centering
  \includegraphics[width=1.0\linewidth]{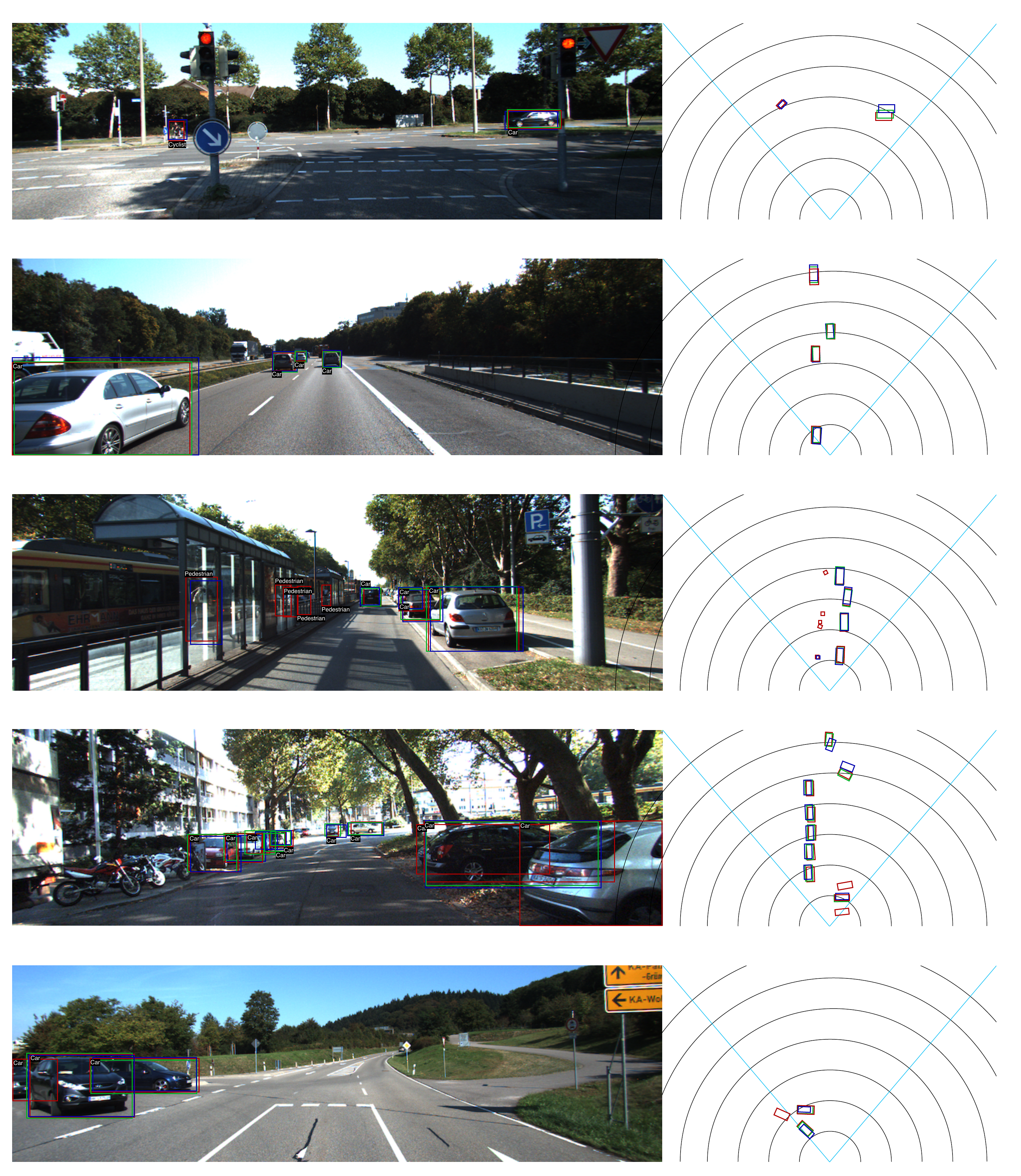}
  \caption{Qualitative comparison on KITTI. Each row shows the image and 3D predictions projected to the image on the left and BEV predictions on the right. Red, blue, and green indicate ground-truth, DD3D, and MonoLiG, respectively.}
  \label{fig:supp-kitti-qual}
\end{figure*}

We present predictions on KITTI dataset from the base DD3D detector and DD3D \cite{park2021dd3d} trained with MonoLiG in \cref{fig:supp-kitti-qual}. We show some of the best cases along with the failure cases. Our method localizes the \textit{Car} class better in the BEV space, and our predictions are closer to the ground-truth boxes compared to the base detector, but for the \textit{Pedestrian} and \textit{Cyclist} class our approach has more false negatives, which the base detector detects but our detector fails.

\end{document}